\documentclass[conference,a4paper]{IEEEtran}

\usepackage[pagebackref=false,breaklinks=true,letterpaper=true,colorlinks,bookmarks=false]{hyperref}
\usepackage{subfigure}
\usepackage{times}
\usepackage{epsfig}
\usepackage{graphicx}
\usepackage{amssymb}
\usepackage{caption} 
\usepackage{refstyle}
\usepackage{amsmath}
\usepackage{tabularx}
\usepackage[dvipsnames]{xcolor}
\usepackage{algorithm}
\usepackage[noend]{algpseudocode}
\usepackage{enumerate}

\newcommand{\lit}{literature}
\newcommand{\ad}{autonomous driving}

\begin{document}
%
\title{Road Detection through Supervised Classification}

\author{*Yasamin Alkhorshid, **Kamelia Aryafar, *Sven Bauer, and *Gerd Wanielik\\
\emph{*Professorship of Communications Engineering,Chemnitz University of Technology}\\ 
\emph{**Computer Science Department, Drexel University}\\
\emph{{\{yasamin.alkhorshid, sven.bauer, gerd.wanielik\}}@etit.tu-chemnitz.de,}
\emph{kca26@drexel.edu}
}

\maketitle

\begin{abstract}
Autonomous driving is a rapidly evolving technology. Autonomous vehicles are capable of sensing their environment and navigating without human input through sensory information such as radar, lidar, GNSS, vehicle odometry, and computer vision. This sensory input provides a rich dataset that can be used in combination with machine learning models to tackle multiple problems in supervised settings. In this paper we focus on road detection through gray-scale images as the sole sensory input. Our contributions are twofold: first, we introduce an annotated dataset of urban roads for machine learning tasks; second, we introduce a road detection framework on this dataset through supervised classification and hand-crafted feature vectors.

\end{abstract}


\IEEEpeerreviewmaketitle

\section{Introduction}

Autonomous driving has been among the most attractive research topics in the road safety applications~\cite{geiger2012we,urmson2008autonomous, shim2015autonomous,jo2015development}. Self-driving cars have been increasingly tested in various road scenarios from highways to urban roads. Autonomous cars provide many benefits such as avoiding traffic collisions, increasing roadway capacity, higher speed limits and reduced involvement of occupants with navigation and driving. These lucrative benefits have spurred interest in advanced driver assistance systems (ADAS). 

Autonomous vehicles detect surroundings using different sensory inputs such as radar, lidar, GNSS, Odometry, and computer vision. Among all these sensors, camera has been the most affordable sensor which is suitable for detection of different types of objects such as pedestrians, other vehicles on the road or lane marking detection. Due to the existence of many potential applications, employing visual perception for semantic understanding of the road scenes has been widely studied in the industry and academia~\cite{Alkhorshid2015LMD}. On the other hand, the use of camera as the primary sensory input is entangled with the high complexity due to many active disturbances and noise and the variety of moving objects surrounding as well as ego-vehicle movement during the processing. 

\begin{figure}[t!]
\centering
\includegraphics[scale= 0.37]{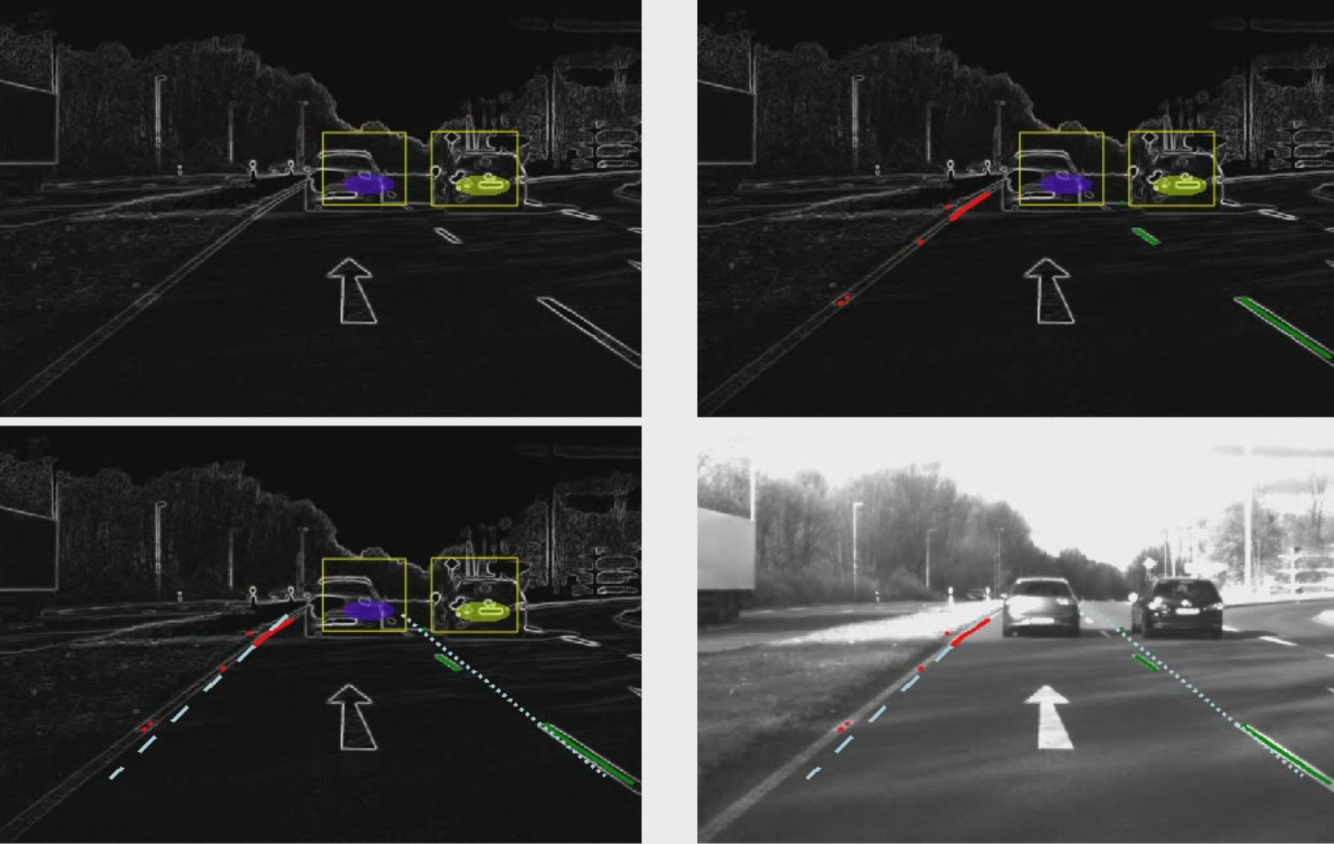}
\vspace{5mm}
\caption{{Conventional Road / Lane marker detection and road
 detection in ADAS is illustrated \cite{Alkhorshid2015LMD} on the benchmark dataset.}}
\label{fig:sample}
\end{figure}

One of the main steps in developing a production-grade autonomous vehicle is a robust road and lane marking detection. Figure~\ref{fig:sample} illustrates a sample lane marking detection and road detection approach. Detection and tracking of lane marking is essential for driving safety and intelligent vehicle~\cite{Alkhorshid2015LMD,Hillel2014recentProgress}. Offline road understanding and Lane detection algorithms are generally composed of multiple modules: image pre-processing, feature extraction and model-fitting~\cite{Son2014illumination,Batista2014perspective,Ding2014urbanIPM,Bottazi2013LMDseg&track}. 

Low-level feature extraction (feature level processing) in every single frame is usually not practical in real-time scenarios due to complexity issues. As an example, calculation of the gradient of the intensity value of ROI in the image easily could get impaired by the existence of other objects or regions with higher contrast than lane markers in the road \cite{Li2014BoundaryLowLevel}. These complexities inspire machine learning methods where a model can be obtained and trained offline and applied in real-time systems. 

A typical machine learning approach to road detection would entail training a classifier on annotated training data and online classification of input image pixels as road~\footnote{In a simple scenario we can assume a binary classification settings where we are only interested in classification of pixels to road or not road.} pixels. A requirement of such road detection or lane marking detection system is the availability of an annotated training dataset. While many datasets exist for various road scenarios, a \textit{publicly available} annotated dataset with multiple sensory sources in urban scenarios is still missing. In this paper we aim to close the gap on developing such dataset. First, we introduce our annotated benchmark dataset in urban roads. Next, we introduce our initial findings on supervised road classification on this dataset using gray scale images as the primary source of information.

 This paper is organized as follows; in Section~\ref{sec:related} we briefly explore the related work on road detection and available datasets for these applications. We introduce our annotated benchmark dataset in Section~\ref{sec:dataset}. We examine a supervised classification approach to road area detection on this dataset and present our initial findings in Section~\ref{sec:method}. Finally we conclude this paper in Section~\ref{sec:conclusion} and propose future research directions.

\section{Related works and challenges}

\label{sec:related}
The accurate perception of the road area is a very important first step for autonomous driving. The road detection has been performed through a various set of sensors both in single modality and with sensor fusions in the \lit\ of computer vision in different road scenarios~\cite{byun2015drivable, fernandez2015comparative, zhou2015efficient, abbas2016novel}. In this paper, we mainly focus on the camera as the sole source of information. 

Various studies in the \lit\ have used camera, i.e. RGB or gray scale images as the sensory input for road, lane marking and sign detection~\cite{alvarez2009vision,he2004color}. These approaches often extract a set of frame based image features and combine them with various machine learning techniques to perform road detection. The extracted features are often categorized as point detectors such as SIFT features, background subtraction and segmentation. Once the features are extracted clustering (unsupervised models) or classification methods are often applied for road detection and tracking. 

The supervised learning models often require partially annotated datasets to train a robust road model. Among all labeling schemes, semantic segmentation of objects in image is the most suitable and efficient input for the learning system. Creating an accurate dataset containing annotation of each meaningful object in each single frame by a human is an expensive and time consuming task. However, some research groups focused on sophisticated methods such as convolutional neural networks to accomplish such a task \cite{Long2014SemanticSeg}.\\
Other research groups develop and improve learning methods based on the samples from the man-made accurate dataset. The most recent published Daimler dataset demonstrates dense pixel semantic annotation for stereo images, capturing the information from 50 cities in germany \cite{CityScapes-Diamler2015}. KITTI dataset equally contains pixel level annotation of images and other sensors information \cite{KITTI-Fritsch2013}. However, Public access for most datasets are limited, for instance KITTI has made a quarter of its dataset available for public. Besides the distinction between the representation of sensory input for creating the database, each research group have their own definition for labeling the object on the image.\\

This limitations motivate a dataset with a standard specifications and the availability for public to customize it depending on their own requirements. The simplicity and ease of use and the ability to change the annotations detail such as the boundaries and the class information of each object are the main advantages of such dataset. In this paper we introduce such benchmark annotated dataset. The annotation is performed on the nominated frames from the video sequences collected from different road scenarios. We believe that a reasonable-size labeled dataset containing samples from diverse road scenarios is a vital input for any supervised learning technique. Later we introduce our proposed machine learning framework for road detection on this dataset.

\section{Creating a Benchmark Dataset}
\label{sec:dataset}

\begin{figure}
\centering
\subfigure{%
    \includegraphics[width=0.2\textwidth]{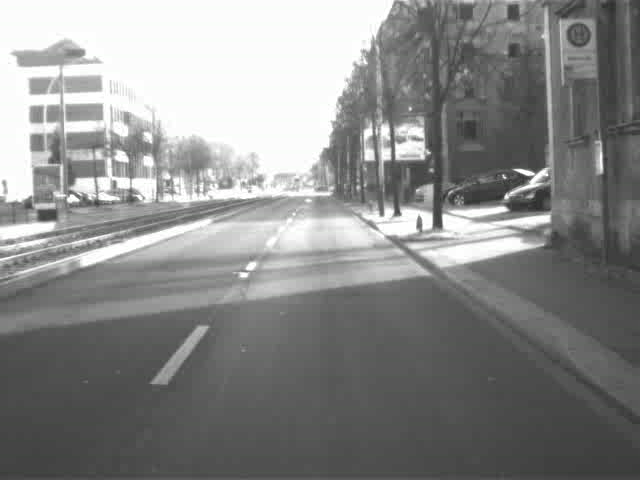}
    \label{spectrum}
}
\subfigure{%
    \includegraphics[width=0.2\textwidth]{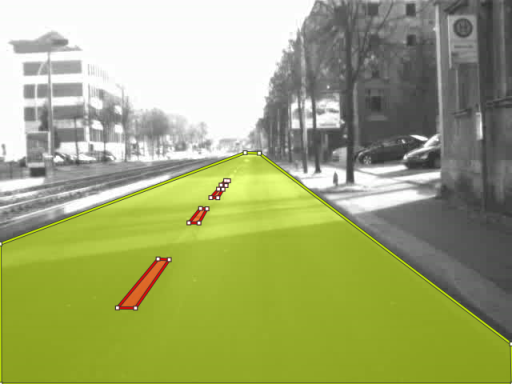}
    \label{mfccs}
}
\subfigure{%
    \includegraphics[width=0.2\textwidth]{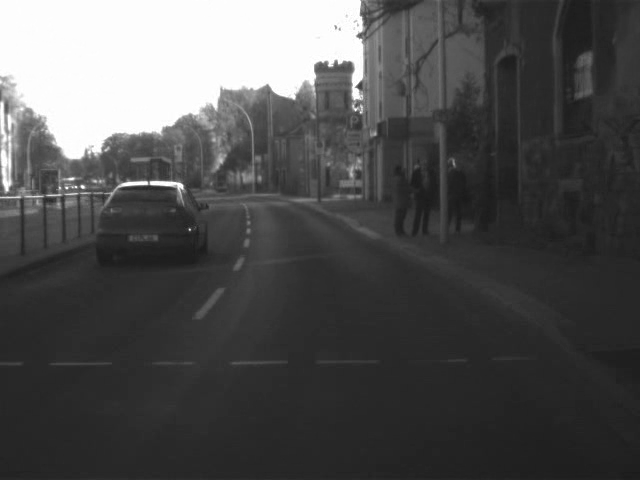}
    \label{spectrum}
}
\subfigure{%
    \includegraphics[width=0.2\textwidth]{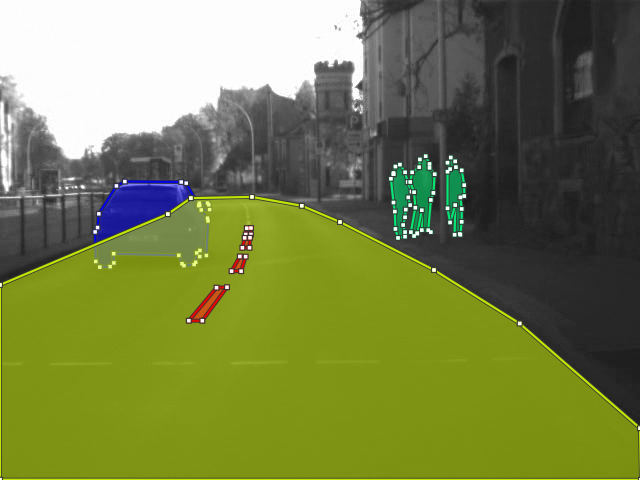}
    \label{spectrum}
}
\subfigure{%
    \includegraphics[width=0.2\textwidth]{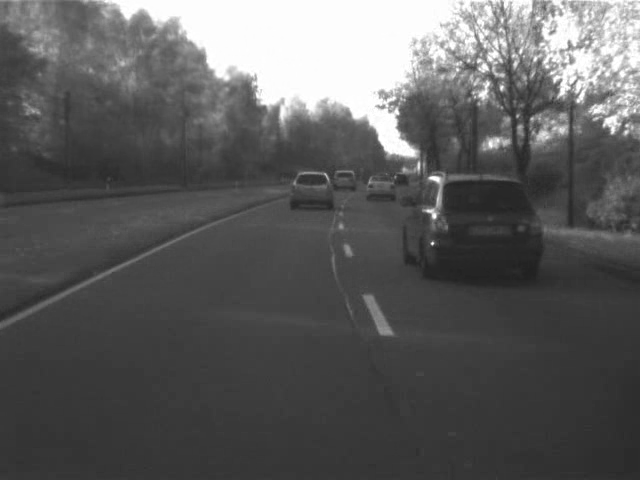}
    \label{spectrum}
}
\subfigure{%
    \includegraphics[width=0.2\textwidth]{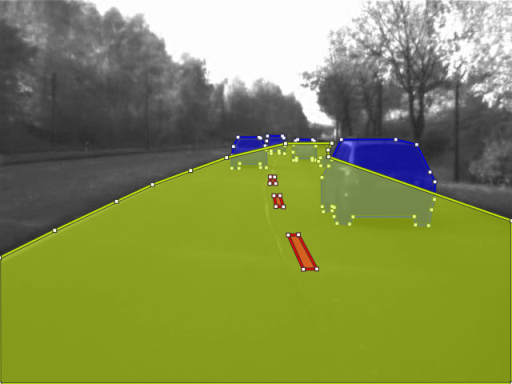}
    \label{spectrum}
}
\vspace{5mm}
\caption{The dataset frames are annotated and labeled into possible four classes: road, lane markers, vehicles, and pedestrians.}
\label{fig.LabeledImage}
\end{figure}

Creating an accurate dataset is a crucial task for a machine learning model of road detection.
Before designing a dataset, various points should be considered in semantic labeling of the dataset.
The first criterion is the type of visual sensor.
As mentioned before, there are several labeled datasets for the stereo and color camera.
The second criterion is the specifications of the dataset which determines the type of objects which we are willing to train our learning algorithm on.

We define classes to which each object belongs to and define a set of attributes associated with each class.
For example, a road class always contains lane markers. The lane markers can also be divided into two sub classes: dashed lane markers and continuous lane marker. This provide a precise description of an object in the scene and its associated attributes.
The joint representation of object class and attributes can help the learning algorithm to interpret the information for object recognition and prediction more precisely \cite{semanticSeg-zheng2014}; since for instance in the example we mentioned before, in case of recognizing the road boundaries correctly, the algorithm will be looking for lane markers only inside the road regions and thus helps in increasing the detection rate for lane marking class.
Another criterion is to have a set of diverse samples from different road scene and weather condition and clarity and illumination.
The higher the number of diverse samples, the more we have enriched training set with variety of street condition and increases the accuracy of detection. To this end, we consider the frames in sunny, illumination variation on the road, shadow, empty road, traffic, crossing, joining road, roads with tram line in the middle, under the bridge scenarios, and autobahn.

Considering all the aforementioned standards for creating benchmark dataset, we setup our unique benchmark dataset consisting of more than 6000 minutes test-drive in German urban and autobahn.
Since annotating this huge information is tedious and expensive, we selected limited number of frames from 90 minutes video sequences for labeling process.
The captured gray scale images were augmented with the class specifications we mentioned earlier.
The sequences captured from a forward looking monocular camera mounted on the front windshield of our test vehicle Carai-1 \cite{schubert2010Carai}.
Sequences were captured with the frequency of 30 fps and $640\times480$ gray scale pixel format.
Layer Annotation tool \cite{layerAnnotation} employed to create our human annotated ground-truth database.
500 frames are color labeled into four object class of interest: road, lane markers, pedestrian, and cars Figure~\ref{fig.LabeledImage}.
Having the annotation information in an \textit{xml} file helped us to employ the data in our algorithm more quickly and reduce the processing complexity and power consumption.
We intend to increase the size and diversity of our database and include other sensory inputs acquired at the same time of capturing the image data.\\

\section{Road Area Classification using the Proposed Dataset}
\label{sec:method}

\subsection{Classification Scheme}
\label{sec:scheme}
To demonstrate the use of the dataset we implemented a binary classifier that shall distinguish between road and non-road areas. This is an information that, for instance, can be used to improve the performance of lane and lane marking detection algorithms by excluding areas that otherwise introduce a lot of clutter line detections. Our classification scheme employs a cascade of AdaBoost classifiers. Binary decision trees with a tree depth of two levels were used as week learners. 

The feature values of the feature vector used by the binary decision trees are solely based upon the gray scale image frames. We added the pixel values, a histogram of those pixel values, and per pixel gradient directions and gradient magnitudes of the region of interest (ROI) to the feature vector. We intended to keep the feature vector calculation as simple as possible; consequently, applying the classifier to a dense sliding winding across the entire frame is still reasonably fast. The pixel values of the gray scale image is the representation of the brightness of that pixel which is a number between 0.0 and 1.0. The frequency distribution of pixel values in the ROI is calculated as histogram values with 256 bins. The gradient values and gradient directions are also computed and included without any further quantization. As a consequence, the dimension of the feature vector $x$ mainly depends on the width $w$ and height $h$  of the region of interest:
\[
\text{dim}\,x = 3 \times w \times h + 256
\]

\begin{figure}
\centering
\subfigure{%
    \includegraphics[width=0.21\textwidth]{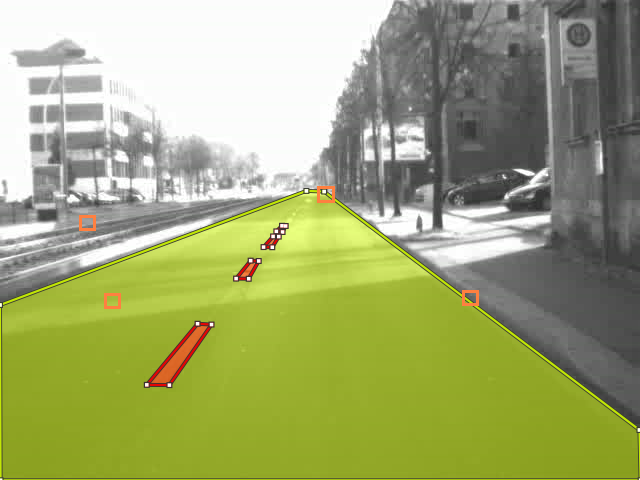}
    \label{spectrum}
}
\subfigure{%
    \includegraphics[width=0.21\textwidth]{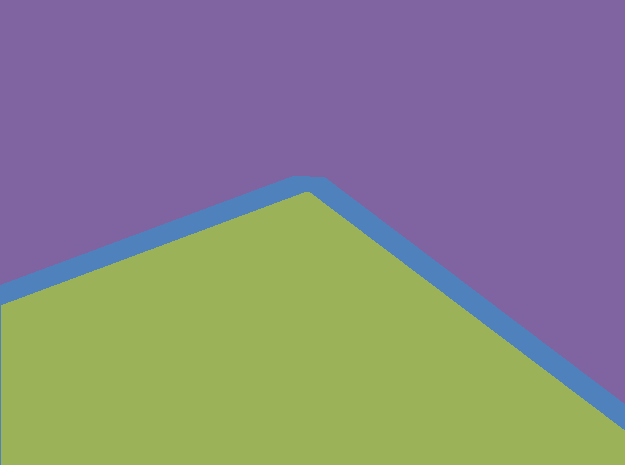}
    \label{spectrum}
}
\subfigure{%
    \includegraphics[width=0.21\textwidth]{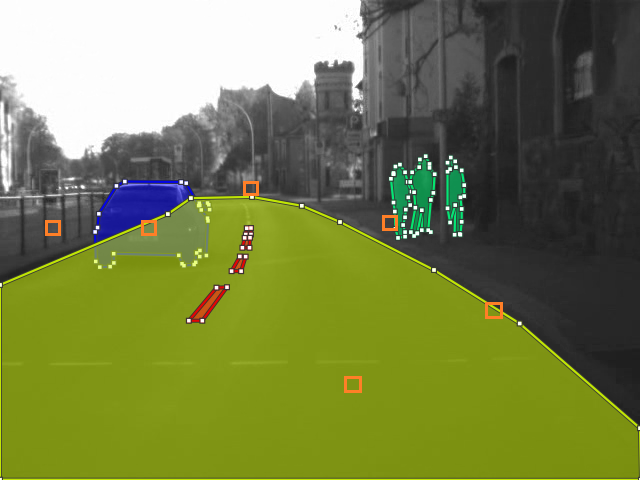}
    \label{spectrum}
}
\subfigure{%
    \includegraphics[width=0.21\textwidth]{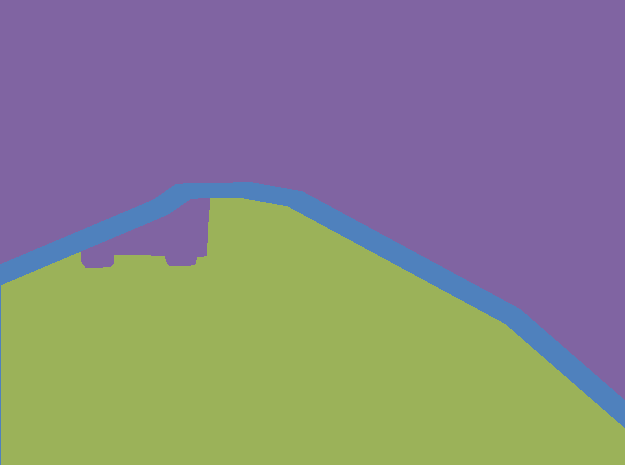}
    \label{spectrum}
}
\subfigure{%
    \includegraphics[width=0.21\textwidth]{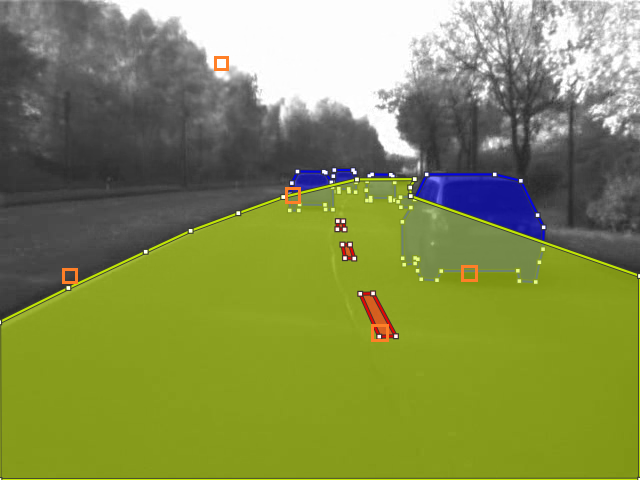}
    \label{spectrum}
}
\subfigure{%
    \includegraphics[width=0.21\textwidth]{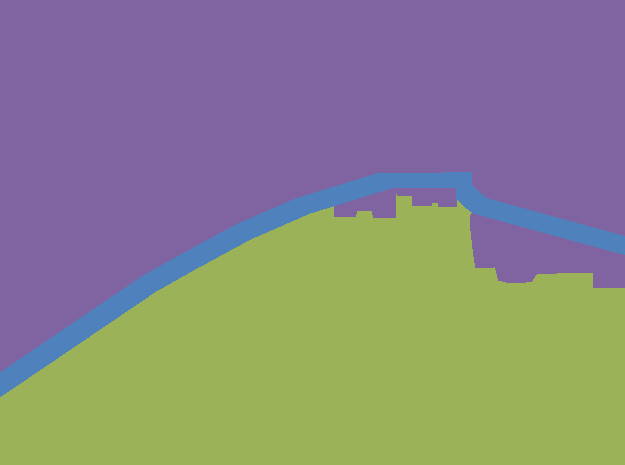}
    \label{spectrum}
}
\vspace{5mm}
\caption{Sliding window ROI classification results for three different frames of the training set. The green areas are regions classified as positive road-areas, blue the mixed areas at the road boundaries and purple the negative regions for non-road areas.}
\label{fig.roiclasses}
\end{figure}

Training this classifier is a two stage process. To begin training the first cascade stage, an initial example set of feature vectors for the positive and negative classes is required. The positive class denotes road areas and the negative class is the representation of non-road areas. For each additional cascade stage, the negative examples correctly rejected by previous stages need to be replaced again requiring the classification of ROIs. In order to distinguish between two classes, the feature vector computation uses the label annotation information. It tests each ROI if it intersects with either the road polygon or any of the car polygons of that frame. This results in the following scenarios:
\begin{itemize}
\item ROI is fully outside of the road polygon.
\item ROI is partially outside of the road polygon.
\item ROI is fully inside of the road polygon and \begin{itemize} \item intersecting a car polygon. \item not intersecting any cars polygon. \end{itemize}
\end{itemize}

A ROI is classified as a negative example if it is fully outside of the road polygon or if it is intersecting a cars polygon. If it is only partially outside of the road polygon, then it is considered as mixed case that could be attributed to either class. By reason of the binary classification nature of the used classifier, we decided to attribute them to the negative class. This leaves only ROIs which are fully inside of the road polygon and which are not intersected by any cars polygon as positive examples. These possible classification cases are shown in Figure~\ref{fig.roiclasses}.

\begin{figure}
\centering
\includegraphics[width=8.8cm]{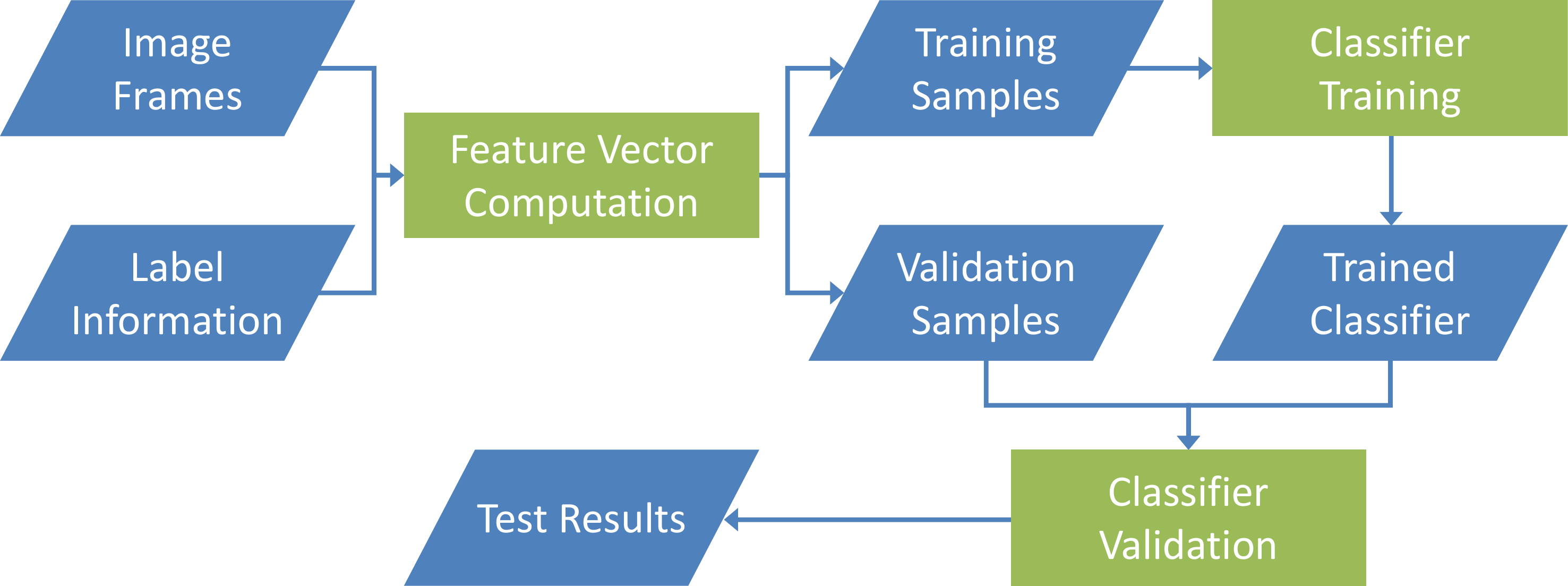}
\vspace{5mm}
\caption{The feature values used by the classifier are a combination of the frames pixel data and the dataset labels for positive/negative classification. Independent samples are then used to validate the trained classifier.}
\label{fig.FlowChart}
\end{figure}

Using this approach, the feature vector computation can automatically create examples for both classes using the proposed dataset. To ensure an independent validation of the trained classifier, only 285 of the so far 500 dataset frames have been used during training. The others 215 frames were then used after training to create positive and negative samples to get the final test results as shown in Figure~\ref{fig.FlowChart} which depicts the overall training and validation process.

\subsection{Training and Evaluation}
  
The cascade AdaBoost is implemented on top of binary decision trees which are employed as weak learners. Assuming N cascades with $\text{DR}_i$ and $\text{FPR}_i $ for detection rate and false positive rate for each layer respectively; the overall detection rate $\text{DR}_t$ and false positive rate $\text{FPR}_t$ of AdaBoost cascade classifier can be computed as follow:

 $\text{DR}_t = \prod_{i=1}^N{\text{DR}_i} $ ,\\
 
 $\text{FPR}_t = \prod_{i=1}^N{\text{FPR}_i} $ 
 
This method suggests that for further reduction of false detection, multiple classifiers should be trained using different sets of negative examples, each giving a cascade stage. Hence, the feature vector needs to come through all stages of the cascade to gain a pass in the testing phase.

The aim of training the classifier is to constitute the road and non-road classes for further evaluation and test. Training of cascade classifier is performed such as following:

\begin{algorithm}[b]
\caption{Adaboost minimizing the weighted error}
\begin{algorithmic}[1]
\Procedure{weak classifier}{$h_t(x)$ }
\vspace{2mm}
\State $ f(x) = \sum_{t=1}^T \alpha_t h_t(x) $
\vspace{1mm}
\State $ H(x)=sign(f(x)) $ \Comment the final classifier \\
\vspace{1mm}
\For{ $(x_1, y_1), \ldots, (x_m, y_m);
 x_i \in \mathcal{X}, y_i \in \{0, +1\}$} 

 \State  $W_{1}(i)=1/m$. \Comment Initialize weights

\For{$t=1,...,T $}
\If {$ \epsilon_t \geq 1/2 $}  
\State stop

\EndIf
\EndFor
\State \textbf{Set} $ \alpha_t = \frac{1}{2}\log(\frac{1 - \epsilon_t}{\epsilon_t}) $
\State $H(x)=sign\left( \sum_{t=1}^{T}\alpha_{t}h_{t}(x)\right)$ \Comment  Output the final classifier
\EndFor
\EndProcedure
\end{algorithmic}
\label{alg:adaboost}
\end{algorithm}
\vspace{5mm}

\begin{enumerate}[(a)]
\item Initial positive and negative class-sets are created by randomly collecting 140 samples from each labeled frame. For this purpose, two uniform random values within the image dimension range are drawn to get the position of a $15\times15$ pixel region of interest. This region is classified as positive or negative as described in section \ref{sec:scheme} and saved accordingly. This procedure is repeated until the number of desired samples for each class of the frame is collected. As 285 frames were used, the total example set consisted of almost 40.000 examples for each class.
\item Having positive and negative samples created, the first cascade classifier is trained. The training process is done by training binary decision trees (BDT) as weak learners for the AdaBoost (see Algorithm \ref{alg:adaboost}). Initially for the first BDT the samples are equally weighted. The false positive and false negative rates are enumerated and the classification results are used to compute the weak learners weight within the AdaBoost voting. Furthermore, for the following BDTs trainings the weights of correctly classified examples are reduced. The threshold for the AdaBoost classifiers weak learner votes is then decreased until it has an acceptable detection rate. More BDTs are trained as long as the false positive rate is too high.
\item After training a cascade stage, correctly rejected negative examples are removed and the new negative samples are sampled from the dataset to replace them. During this phase, the random sampling is replaced with a sliding window approach. 
\item We iterate through step "b" until the detection rate (DR) is still above the set threshold, and continue with step "b" to train another cascade stage.
\end{enumerate}

\vspace{5mm}

\subsection{Evaluation Results}
 \label{sec:results}
 
\begin{figure}[b]
\centering
\includegraphics[scale=1.0]{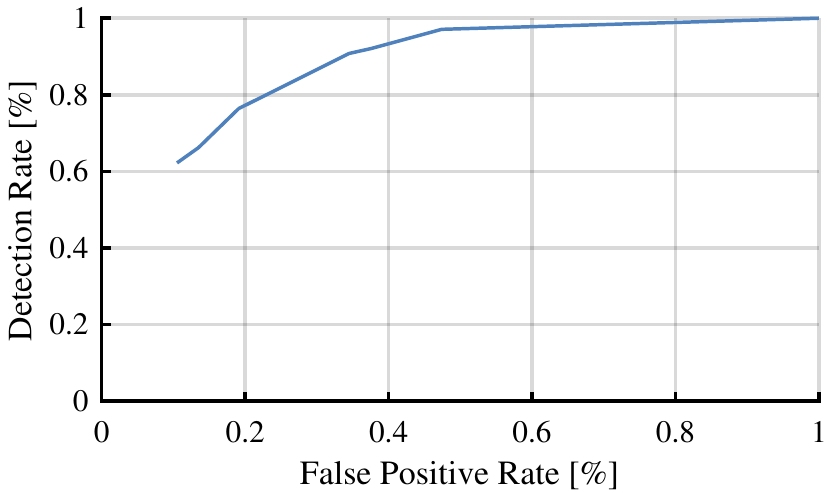}
\vspace{5mm}
\caption{Receiver Operating Characteristic curve (ROC) for evaluating the performance of our binary classifier and to select the best possible optimal model of the test.}
\label{fig.ROC-Chart}
\end{figure}

\begin{figure}[t]
\subfigure{%
    \includegraphics[width=0.231\textwidth]{./Image1.png}
    \label{spectrum}
}
\subfigure{%
    \includegraphics[width=0.231\textwidth]{./Image3.png}
    \label{spectrum}
}
\subfigure{%
    \includegraphics[width=0.231\textwidth]{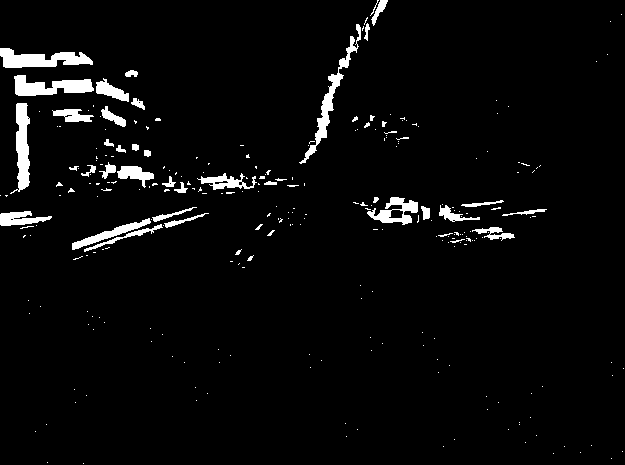}
    \label{spectrum}
}
\subfigure{%
    \includegraphics[width=0.231\textwidth]{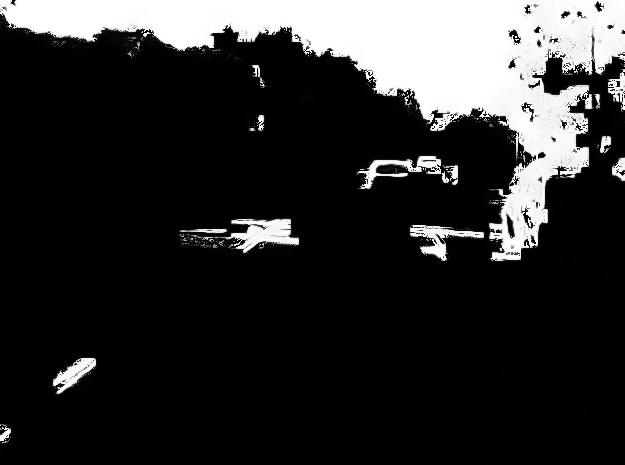}
    \label{spectrum}
}
\subfigure{%
    \includegraphics[width=0.231\textwidth]{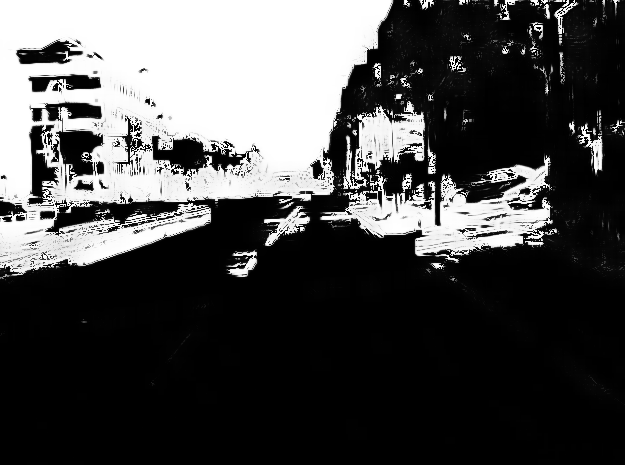}
    \label{spectrum}
}
\subfigure{%
    \includegraphics[width=0.231\textwidth]{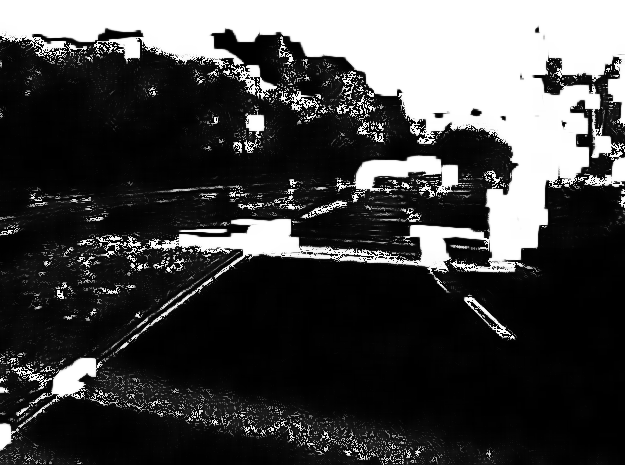}
    \label{spectrum}
}
\subfigure{%
    \includegraphics[width=0.231\textwidth]{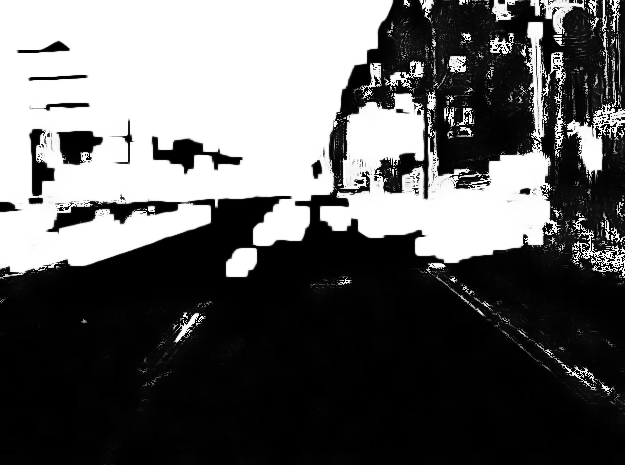}
    \label{mfccs}
}
\subfigure{%
    \includegraphics[width=0.231\textwidth]{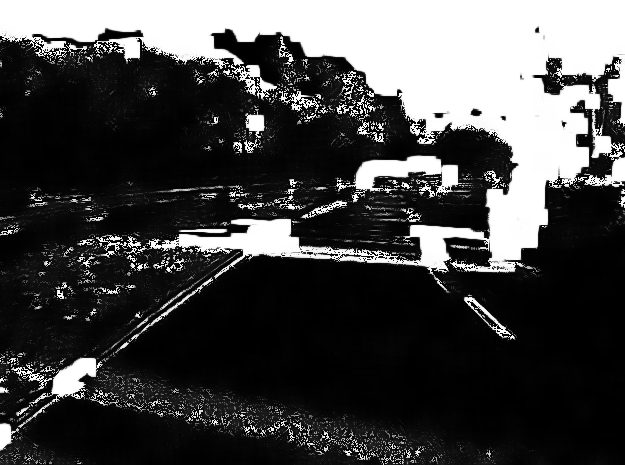}
    \label{mfccs}
}
\subfigure{%
    \includegraphics[width=0.231\textwidth]{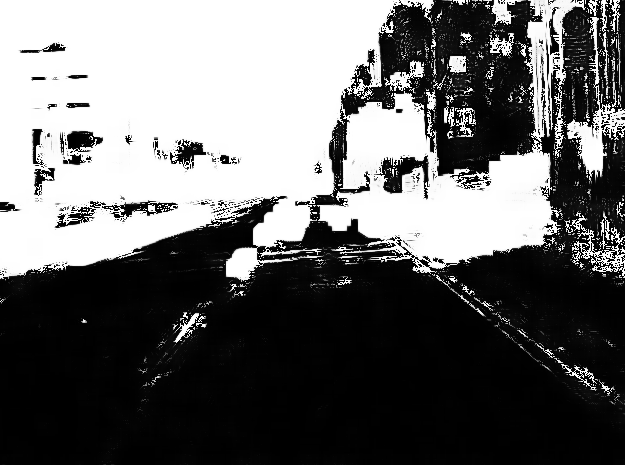}
    \label{spectrum}
}
\subfigure{%
    \includegraphics[width=0.231\textwidth]{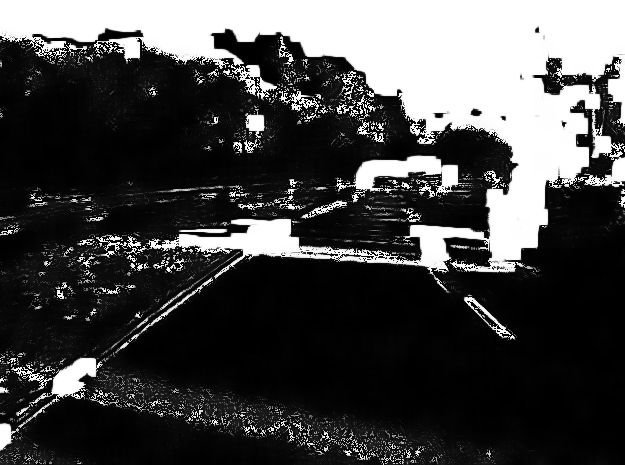}
    \label{spectrum}
}
\vspace{5mm}
\caption{classification Images for each adaboost thresholds.}
\label{fig.ClassificationImage}
\end{figure}

For validation of the classifier, to avoid possible overfitting, positive and negative examples were randomly generated from the section of the dataset not used during training which yielded the following numbers. The evaluation of training process demonstrated the optimal results when the sliding window size is $15\times15$ pixel. Depending on the detection rate, a different false positive rate (FPR) occures, as can be seen in ROC curve in Figure \ref{fig.ROC-Chart}. ROC curve enables us to adjust the optimal detection rate and false positive rate for the trainer.
Conventionally, the right working point is a trade off between DR and FPR. 
As it can be seen from the ROC chart, the final results for each cascade stage illustrates the convergence of adaboost output to the logarithm of likelihood ratio. We obtained a maximum DR of $\mathbf{78\%}$ and $\mathbf{19.8\%}$ FPR.
The results for the validation of the trained classifier against the selected sample are illustrated in Figure~\ref{fig.ClassificationImage}.  The positive road detection results are colored in black and the non-road regions in white.
 
Our initial results can serve as a proof of concept that gray scale images can be used as the sole sensory input for road detection using simple feature vectors and machine learning models. It should be noted that these features are by no means the best feature vectors for road detection and have only been selected as the proof of concept. In future studies we will consider the use of more robust feature vectors in combination with other machine learning models to achieve higher detection accuracy and generalization. 

\section{Conclusion and discussion}
\label{sec:conclusion}

This paper represents a new annotated benchmark dataset of urban roads. Our contributions were twofold; first we introduced the benchmark dataset which is annotated manually by semantic segmentation of frames. Next, we demonstrated a supervised learning approach as a simple proof of concept on the benchmark dataset for road detection as the primary application. To this end, we extracted a set of low-level image features from each gray scale frame. We then used a cascade of adaboost binary decision tree as the supervised learning model for classification of road pixels. Our initial results illustrated that simple supervised classification methods in combination with gray scale image features can be applied for the automated driving challenging tasks such as road detection and scene understanding. \\

The future research on this dataset will be on three distinct areas; the proper selection of informative feature vectors, the selection of a scalable learning models and the combination of other data modalities such as odometery information of the vehicle and other on-board sensors with the hope of increased detection accuracy. We anticipate that an adaptive method with a pre-calculated road model can enhance the robustness of the detection for challenging road detection scenarios such as urban roads. We hope that this dataset can serve as public benchmark to further the active research on road detection and \ad.

\IEEEtriggeratref{10}
\IEEEtriggercmd{\enlargethispage{-4.52in}}


%

{\small
\bibliographystyle{unsrt}
\bibliography{ICPR}
}

\end{document}